\documentclass[letterpaper,twocolumn,10pt]{article}
\usepackage{usenix}

\usepackage{tikz}

\usepackage{filecontents}

\usepackage{cite}
\usepackage{amsmath,amssymb,amsfonts}
\usepackage{graphicx}
\usepackage{textcomp}
\usepackage{xcolor}
\usepackage{subcaption}

\def\BibTeX{{\rm B\kern-.05em{\sc i\kern-.025em b}\kern-.08em
    T\kern-.1667em\lower.7ex\hbox{E}\kern-.125emX}}

\usepackage{balance}
\usepackage{mathrsfs}
\usepackage{bm}
\usepackage{svg}
\usepackage{tabularx}  
\usepackage{booktabs}  
\usepackage{algpseudocode}
\usepackage{algorithm}
\usepackage{multirow}

\usepackage{comment}
\usepackage[textwidth=1.5cm,textsize=scriptsize]{todonotes}

\newcommand{\new}[1]{\textcolor{black}{#1}}

\RequirePackage{amsmath}
    
    \DeclareMathOperator*{\argmin}{arg\,min}

\begin{document}

\date{}

\title{\Large \bf Moshi Moshi? A Model Selection Hijacking Adversarial Attack}

\author{
{\rm Riccardo Petrucci}\\
University of Padua, Italy
\and
{\rm Luca Pajola}\\
University of Padua, Italy
\and
{\rm Francesco Marchiori}\\
University of Padua, Italy
\and
{\rm Luca Pasa}\\
University of Padua, Italy
\and
{\rm Mauro Conti}\\
University of Padua, Italy
} 

\maketitle

\begin{abstract}
\new{
Model selection is a fundamental task in Machine Learning~(ML), focusing on selecting the most suitable model from a pool of candidates by evaluating their performance on specific metrics.
This process ensures optimal performance, computational efficiency, and adaptability to diverse tasks and environments.
Despite its critical role, its security from the perspective of adversarial ML remains unexplored.
This risk is heightened in the Machine-Learning-as-a-Service model, where users delegate the training phase and the model selection process to third-party providers, supplying data and training strategies.
Therefore, attacks on model selection could harm both the user and the provider, undermining model performance and driving up operational costs.
}

\new{
In this work, we present \textbf{MOSHI} (\textbf{MO}del \textbf{S}election \textbf{HI}jacking adversarial attack), the first adversarial attack specifically targeting model selection.
Our novel approach manipulates model selection data to favor the adversary, even without prior knowledge of the system.
Utilizing a framework based on Variational Auto Encoders, we provide evidence that an attacker can induce inefficiencies in ML deployment.
We test our attack on diverse computer vision and speech recognition benchmark tasks and different settings, obtaining an average attack success rate of 75.42\%. 
In particular, our attack causes an average 88.30\% decrease in generalization capabilities, an 83.33\% increase in latency, and an increase of up to 105.85\% in energy consumption.
These results highlight the significant vulnerabilities in model selection processes and their potential impact on real-world applications.
}
\end{abstract}

\section{Introduction}

Adversarial Machine Learning (AML) represents a rapidly advancing subfield within Artificial Intelligence (AI) and cybersecurity, focusing on studying and mitigating attacks that exploit vulnerabilities in Machine Learning (ML) models.
The seminal work by Barreno et al. in 2006, titled ``\textit{Can machine learning be secure?}''~\cite{barreno2006can}, marked the inception of this critical area of research.
This pioneering study introduced a taxonomy of attacks based on three key properties: security violation, influence, and specificity.
Since then, extensive research has been conducted to explore various aspects of AI security, addressing numerous challenges and significantly enhancing our understanding of safeguarding ML systems against adversarial threats.
\par
Adversarial attacks can be broadly categorized into several families, each targeting different aspects of ML systems
For instance, the \textit{evasion attack} involves manipulating input data to deceive a model into making incorrect predictions~\cite{biggio2013evasion}.
This attack exploits vulnerabilities in the model's decision boundary to achieve misclassification.
\textit{Model poisoning} refers to the deliberate contamination of training data to corrupt the learning process~\cite{biggio2012poisoning}.
This attack aims to degrade the model's performance or manipulate its predictions by introducing maliciously crafted data in the training set during the training phase.
The \textit{membership attack} aims to determine whether a specific data point was part of the model's training dataset~\cite{shokri2017membership}.
This attack exploits the model's overfitting tendencies to infer the presence of individual data samples, potentially compromising data privacy.
\par
With the growing integration of AI into commercial and industrial assets, it is becoming increasingly vital to understand how adversarial attacks can be executed in real-world applications and to evaluate their potential impact.
An example of a popular AI incident is Microsoft's AI chat bot ``Tay''~\cite{wolf2017why}. 
Tay was a chatbot launched on Twitter that was intended to learn from user interactions.
Unfortunately, due to the actions of malicious users, Tay was swiftly influenced to adopt offensive and racist language.
\par
Exploring various attack vectors that could compromise the quality of ML development and deployment is paramount.
\new{However, despite its critical role, \textit{no prior work in the literature has considered the possibility of adversarial attacks targeting the model selection phase}.}
Model selection, the process of choosing the most suitable algorithm or model from a pool of candidates for a specific task, is essential for ensuring AI systems' effectiveness and generalization capability.
This issue is especially pressing in the current era, where the Machine-Learning-as-a-Service (MLaaS) paradigm is widely adopted.
\new{
If extended to target the model selection phase, adversarial attacks could have far-reaching consequences, degrading model performance and significantly impacting energy consumption and execution costs.}
Such attacks could force models to use more computationally expensive algorithms, leading to higher operational costs for both the user and the service provider.
This raises critical challenges, as the increased resource consumption, prolonged task execution, and degraded performance would burden both parties, resulting in elevated costs in running ML models.
Therefore, in this article, we address the critical question: \textit{can an attacker influence the model selection phase?}
\paragraph{Contribution.}
\new{
In this paper, we are the first to present an adversarial attack against the model selection phase.
Our attack called \textbf{MOSHI} (\textbf{MO}del \textbf{S}election \textbf{HI}jacking adversarial attack) targets only the validation set of an ML system, manipulating the model selection process to favor a model with properties advantageous to the attacker.
We generate adversarial samples using a modified conditional Variational Autoencoder (VAE) that optimizes a hijack metric while leaving the victim’s original loss metric and code unaltered.
The result is a scenario where the model that best aligns with the attacker’s hijack metric also appears to minimize the validation loss, ultimately being selected by the model selection process.
Our contributions can be summarized as follows.
}
\begin{itemize}
    \item \new{We introduce MOSHI, the first adversarial attack for model selection hijacking.
    Our attack leverages a VAE optimizing a specifically crafted hijack metric to inject malicious samples in the validation set of an ML system. In our experiments, we demonstrate that MOSHI can undermine critical aspects of ML performance like generalization capability, latency, and energy consumption.}
    \item \new{Our attack methodology considers various scenarios based on the attacker’s level of knowledge, namely \textit{white-box} and \textit{black-box} scenarios.
    These experimental settings demonstrate the effectiveness of our attack and highlight its practical feasibility across a wide range of real-world scenarios.}
    \item \new{We assess the impact of our proposed attack across three distinct benchmarks in computer vision and speech recognition, achieving an average Attack Success Rate (ASR) of 75.42\%.
    Notably, in our evaluation, our attack results in an 88.30\% reduction in generalization capabilities, an 83.33\% increase in latency, and a 105.85\% rise in energy consumption (12.65\% when estimated through a simulator, 167.98\% when estimated through $\ell_0$ norm).
    }
\end{itemize}

\paragraph{Organization.}
\new{
The rest of this paper is organized as follows.
Section~\ref{sec:related} presents an overview of related works in AML, while Section~\ref{sec:background} provides background knowledge on ML systems principles.
We present our system and threat model in Section~\ref{sec:stmodel}, providing motivations and actionable scenarios for our attack.
MOSHI's methodology is detailed in Section~\ref{sec:methodology}, and in Section~\ref{sec:evaluation} we provide our experimental settings and evaluation.
Finally, we discuss our results in Section~\ref{sec:discussion}, and Section~\ref{sec:conclusion} concludes our work.
}
\section{Related Work}
\label{sec:related}

In this section, we review the adversarial attacks from recent years that are related to the MOSHI attack presented in this work.
These include various approaches aimed at influencing model behavior during training or evaluation, providing context for the novelty and scope of MOSHI in comparison to existing methods.

\paragraph{Poisoning Attacks.}
In AML, model poisoning deliberately manipulates the training process by introducing maliciously crafted data into the training dataset~\cite{tian2022comprehensive}.
This manipulation aims to corrupt the model's learning algorithm, leading to compromised performance and erroneous outputs.
Such attacks can subtly alter the decision boundaries, degrade the model's accuracy, or embed specific vulnerabilities, enabling the attacker to influence future predictions or decisions.
Model poisoning exploits the trust in the training data, highlighting the importance of securing and validating data sources in ML workflows.
\new{
Our proposed attack involves poisoning the victim's data but differs significantly from traditional techniques in three key ways: \textit{(i)} MOSHI targets only the validation set, leaving the training set untouched; \textit{(ii)} it does not modify the trained model parameters but manipulates the model selection process to choose a less optimal (yet legitimate) model; and \textit{(iii)} it remains imperceptible, as it introduces no detectable anomalies during training.
This unique design makes MOSHI highly adaptable across various scenarios, as it exclusively impacts validation data without altering any aspect of the victim’s model selection process.
}

\paragraph{Availability Attacks.}
Adversarial attacks targeting availability, often called Denial-of-Service (DoS) attacks, aim to disrupt the regular operation of AI systems, rendering them unusable or significantly degrading performance.
These attacks flood the system with excessive, irrelevant, or malicious data, overwhelming its processing capabilities and leading to resource exhaustion.
This can cause delays, reduced accuracy, or complete shutdowns of the AI services, effectively preventing legitimate users from accessing the system.
Examples of attacks are exploiting software vulnerabilities~\cite{xiao2018security}, Camouflage Attack~\cite{xiao2019seeing}, and Zero-Width Attack~\cite{pajola2021fall}.
\new{
Likewise attacks on the availability, MOSHI can introduce an increased utilization of resources.
However, MOSHI objective function is purely guided by a hijack metric measuring model properties like latency, energy consumption, or generalization error.
}

\paragraph{Energy-Latency Attacks}
During the deployment phase, ML models, particularly Deep Neural Networks (DNNs), exhibit high demands on computational resources and memory capacity.
For this reason, more and more specialized hardware is being developed, such as Big Basin~\cite{hazelwood2018applied}, Project BraiwWave~\cite{chung2018serving}, and Tensor Processing Units (TPUs)~\cite{jouppi2018motivation}.
We refer to these highly specific chipsets as Application-Specific Integrated Circuits (ASIC).
Shumailov et al.~\cite{shumailov2021sponge} kick-started the AML research trend focusing on the increase in consumption of resources (e.g., energy, latency) of victims' models.
They presented the \textit{sponge examples attack}: the attack generates specific adversarial samples, termed sponge samples, which are introduced to the victim's model.
Unlike previously known attacks, the primary objective of this method is to escalate the latency and energy consumption of the target model, disregarding the prediction accuracy of the samples.
Consequently, the primary goal of this attack is to induce an availability breakdown.
The attack can be carried out by working on either \textit{(i)} the computation dimension, which increases the internal representation dimension to increase the time complexity of computations (useful in Natural Language Processing); \textit{(ii)} data activation sparsity, ss ASICs use runtime data sparsity to increase efficiency (like skipping operations), thus inputs that lead to less sparse activations will lead to more operations being computed (useful in Computer Vision).
Cina et al.~\cite{cina2022energy} introduced a poisoning attack aiming to increase the latency and energy consumption of a victim's model, without affecting the prediction performance.
The attack is carried out via sponge training (a type of model poisoning), which is a carefully crafted algorithm that minimizes the empirical risk of the training data while maximizing energy consumption.
The algorithm objective is pursued by increasing the model's activation (i.e. $\ell_0$ norm: the number of firing neurons) to undo the advantages of ASIC accelerators.

\section{Background}
\label{sec:background}

This section describes in depth the basis of supervised learning (Section~\ref{subsec:sl}) and the role of model selection (Section~\ref{subsec:ms}).
We opt for a detailed description aiming to provide readers all the elements to understand the idea behind the attack we propose in this work, the first of its kind. 
This section follows the notation of~\cite{shalev2014understanding, goodfellow2016deep, oneto2020model}.

\subsection{Supervised Learning}
\label{subsec:sl}
Supervised Learning (SL) is an ML paradigm wherein the training process involves presenting a labeled dataset (experience) to a learning algorithm.
This process results in the development of a model capable of assigning labels to input samples.
The paradigm is composed of the following elements: a learning algorithm $\mathscr{A}$; an arbitrary set of sample that we wish to label $\mathcal{X}$; a set of possible labels $\mathcal{Y}$; $\mathfrak{G}: \mathcal{X} \to \mathcal{Y}$, i.e. a rule which assigns labels $y \in \mathcal{Y}$ to input points $\bm{x} \in \mathcal{X}$, also called labeling or target function, generally unknown to the learner; $\mathcal{D}$, a probability distribution over $\mathcal{X}$, unknown to the learner.
In SL, the aim of our learning algorithm or learner $\mathscr{A}$ is to find a hypothesis $h$ that approximate $\mathfrak{G}$, through another rule $h: \mathcal{X} \to \mathcal{Y}$, which maps inputs $\bm{x} \in \mathcal{X}$ into labels $\hat{y} \in \mathcal{Y}$.
\subsubsection{Experience}
This goal is carried out by providing to $\mathscr{A}$ the experience, which in the SL framework is a set of samples $\mathcal{\hat{S}} \subseteq \mathcal{X} \times \mathcal{Y}$.
In the real world, we have a limited number of samples.
Thus we will consider $\mathcal{S} \subset \mathcal{\hat{S}}$ to be a finite set, composed of $n$ samples $(\bm{x}_i, y_i)$, where $y_i = \mathfrak{G}(\bm{x}_i)$, and $\bm{x}_i$ is sampled from $\mathcal{X}$ according to the distribution $\mathcal{D}$ for $1 \leq i \leq n$.
Therefore, we can express our learner as $\mathscr{A}: \mathcal{S} \to \mathcal{H}$, where $\mathcal{H}$ is the Hypothesis Space, which constitutes a set of all functions which can be implemented by the ML system ($h \in \mathcal{H}$).
\subsubsection{Task}
It governs the way the ML algorithm should process each sample $\bm{x} \in \mathcal{X}$, by way of example, remaining in the SL framework, the most common tasks are a \textit{classification} and \textit{regression}.
In classification, $\mathscr{A}$ is asked to find a function $h: \mathcal{X} \to \mathcal{Y}$, specifying which of $|\mathcal{Y}|$ classes
some input $\bm{x} \in \mathcal{X}$ belongs to. 
In regression, $\mathscr{A}$ is asked to find a function $h: \mathcal{X} \to \mathcal{Y}$, which maps inputs $\bm{x} \in \mathcal{X}$ into some numerical value, usually $\mathcal{Y} = \mathbb{R}$, therefore the format of the output is different compared to the classification task.
\subsubsection{Performance Measure}
The capability of $h$ in approximating $\mathfrak{G}$ is evaluated via a quantitative performance measure usually specific to the task being carried out.
In this work 
 we focus on benchmark classification tasks with balanced datasets, thus as a performance measure, we adopted the accuracy, i.e., the proportion of examples for which the model produces incorrect output. 
Assuming that we are dealing with deterministic rules, 
we can introduce the concept of \textit{true error} as the probability that $h$ will fail to classify an instance drawn at random according to $\mathcal{D}$:
\begin{equation}
\mathcal{L}_{\mathcal{D}} (h) = \mathbb{E}_{\bm{x} \sim \mathcal{D}}[\mathcal{L}(h, \bm{x},\mathfrak{G}(\bm{x}))].
\end{equation}
However, a general learning algorithm cannot access the probability distribution $\mathcal{D}$. 
Thus, it is not possible to compute $\mathcal{L}_{\mathcal{D}} (h)$, and we can only use the learner $\mathscr{A}$, with the known training data $\mathcal{S}$, resulting in:
   \begin{equation}
 h : \mathscr{A}(\mathcal{S}).
   \end{equation}
To measure the performance of our model, we can estimate the true error rate by computing the error rate on unseen data. Therefore, we employ a test set of data:
$\mathcal{S}^{Test} = \{(\bm{x}_j,\mathfrak{G}(\bm{x}_j)): 1 \leq j \leq k, \quad \bm{x}_j \in \mathcal{X} \}$ which is separate from the data used for training the ML system: $\mathcal{S}^{Test} \cap \mathcal{S} = \emptyset $.
Thus, we can define the test error as follows:
\begin{equation}
    \mathcal{L}_{Test} (\mathscr{A}(\mathcal{S}), \mathcal{S}^{Test}) = \mathbb{E}_{(\bm{x},y) \in \mathcal{S}^{Test}}[\mathcal{L}(\mathscr{A}(\mathcal{S}), \bm{x},y)].
\end{equation}

\subsection{Model Selection}
\label{subsec:ms}
\new{Model selection involves identifying the best hypothesis $h$ from a set $\mathcal{H}$ with the goal of minimizing true error through hyperparameters tuning.
This section discusses the role of hyperparameters and the process of selecting most suitable configurations.}
\subsubsection{Hyperparameters}
Any hypothesis $h \in \mathcal{H} $ is characterized by a set of hyperparameters (parameters not adjusted by the learning algorithm itself), which delineate the possible set of functions contained in $\mathcal{H}$, on which $\mathscr{A}$ perform the research of the most suitable function. 
Therefore, the selection of the best configuration of hyperparameters $\mathfrak{c}$ in a set of possible configurations $\mathfrak{C}$ is an important problem in learning:
which could be generalized to the problem of choosing between different algorithms $\mathscr{A} \in \mathcal{A}$, each characterized by its configuration of hyperparameters $\mathfrak{c}$.
\begin{equation}
 \mathcal{A}_\mathfrak{C} = \{\mathscr{A}_{\mathfrak{c}} : \mathscr{A} \in \mathcal{A}, \mathfrak{c} \in \mathfrak{C}_{\mathscr{A}} \},
   \end{equation}
where $\mathfrak{C}_{\mathscr{A}}$ is the set of possible configurations of hyperparameters for learner $\mathscr{A}$.
\subsubsection{\new{Selection Process}}
During model selection, the objective is to choose the hypothesis $h$ that minimizes the true error.
Since the true error of a hypothesis cannot be directly computed, we rely on an approximation obtained from a set of unseen data (validation set) drawn from the distribution $\mathcal{D}$. 
This process requires training several models with different hyperparameter configurations $c$ and then approximating the true error for each configuration.
The resulting model with the lowest approximate error will be the candidate chosen to solve the given task.
This procedure is both time and computationally demanding, as the number of possible hyperparameter configurations $c$ is potentially infinite.
A straightforward approach to address this challenge is to use Grid Search.
Grid Search systematically combines and evaluates a predefined set of hyperparameter values.
Although this reduces the complexity by limiting the search space, it can still be computationally expensive, especially when the grid is large or the model is complex.
\subsubsection{Resampling Methods}
Performing model selection is often constrained by the amount of available data.
When data is limited, using a resampling method can be helpful, as splitting the dataset into separate training, validation, and test sets of significant size could be detrimental to the training process.
Resampling techniques allow for more efficient use of the data by repeatedly training and validating the model on different subsets, ensuring better performance evaluation without sacrificing valuable data for training.
Two standard techniques for model selection are the Hold-Out method and k-fold cross-validation.
Both rely on resampling the original dataset $\mathcal{S}$ to create the two essential sets needed for model selection: the training set $\mathcal{S}^{Train}$ and the validation set $\mathcal{S}^{Val}$.
Formally, the training set is used to train the different algorithms $\mathscr{A}_{\mathfrak{c}} \in \mathcal{A}_\mathfrak{C}$, thus obtaining different candidate rules $\mathscr{A}_{\mathfrak{c}}(\mathcal{S}) = h_{\mathfrak{c}}$, then the latter is employed for estimating the true error of each candidate rule, to choose the best one. 
In this work, we focus on the Hold-Out procedure. 
It consists of simply performing resampling on $\mathcal{S}$ without replacement and then using the same $\mathcal{S}^{Train}$ to train all the models, and $\mathcal{S}^{Val}$ to select the best candidate rule as the one with minimum validation error:
\begin{equation} \small
\begin{split}
     \mathcal{L}_{Val} (\mathscr{A}_{\mathfrak{c}}(\mathcal{S}^{Train}), \mathcal{S}^{Val}) = \mathbb{E}_{(\bm{x},y) \in \mathcal{S}^{Val}}[\mathcal{L}(\mathscr{A}_{\mathfrak{c}}(\mathcal{S}^{Train}), \bm{x},y)].
\end{split}
\end{equation}
\section{System and Threat Model}
\label{sec:stmodel}

\new{
In this section, we describe our system and threat model (Section~\ref{subsec:system} and Section~\ref{subsec:threat}) while also providing different motivations and scenarios of importance for our proposed attack (Section~\ref{subsec:motivation}).
}

\subsection{System Model}
\label{subsec:system}

\new{
The systems we consider reflect a typical ML pipeline that addresses diverse real-world tasks, such as image classification and speech recognition.
These pipelines often involve model selection as a critical phase to ensure the generalization capabilities of the model.
On the other hand, this work demonstrates that poisoning model selection can also be used to influence the behaviors and characteristics of an ML algorithm.
Indeed, different domains often impose distinct demands on model characteristics.
For example, real-time applications like autonomous systems require low-latency models to ensure timely decision-making, while tasks that impact human lives or critical infrastructure prioritize accuracy and reliability above all.
Additionally, deployment scenarios further shape these requirements, as energy-efficient models are indispensable for battery-powered devices operating in resource-constrained environments.
}

\subsection{Threat Model}
\label{subsec:threat}

Our proposed attack aims to control the model selection phase by injecting maliciously crafted samples into the sole validation set.  
In this scope, the adversary goal is to let the model selection phase select the model architecture that suits the best target metric, named \textit{hijack metric} $m$. 
The hijack metric is arbitrary and can be chosen by the adversary.
Lastly, we assume that adversaries cannot modify the standard model selection routine, including the loss metric defined by the victim, the underlying code, or the model selection policy.
This means that with MOSHI, we inject samples that influence the loss function evaluation on the validation set, ensuring that the model best suited to the hijack metric achieves a lower loss than other models in the validation grid.
In this work, we explore different types of hijack metrics.

\paragraph{Assumptions.}
Suppose the victim has access to a dataset $\mathcal{S}$, divided in train $\mathcal{S}^{Train}$, validation $\mathcal{S}^{Val}$, and testing $\mathcal{S}^{Test}$ sets.
The former is used to train several model $h_{\mathfrak{c}}$ grouped in a model set identifiable with $\mathfrak{C}$,\footnote{$\mathfrak{c}$ represents the configuration of hyperparameters and $\mathfrak{C}$ the set of all possible configurations of hyperparameters.} the second to perform model selection, and the latter to test the best model performance.
The model selection phase will return the model $h_{\mathfrak{c}^*}$ with the lowest true error estimate on the validation set among those trained:
    \begin{equation}
        \label{best_val}
 h_{\mathfrak{c}^*} = \argmin_{h_{\mathfrak{c}} : \mathfrak{c} \in \mathfrak{C}} \mathcal{L}_{Val}(h_{\mathfrak{c}}, \mathcal{S}^{Val}).
    \end{equation}
In this context, we postulate that the adversary possesses access to the validation set $\mathcal{S}^{Val}$, enabling them to read and modify its data.
Consequently, our newly proposed adversarial family, MOSHI, exhibits characteristics akin to data poisoning, as it allows the attacker to alter the victim’s dataset.
In contrast with the traditional poisoning attack, our proposed attack operates solely at the validation set level, leaving the training set unaltered.

\begin{figure*}[!htpb]
    \centering
    \includegraphics[width=0.65\linewidth]{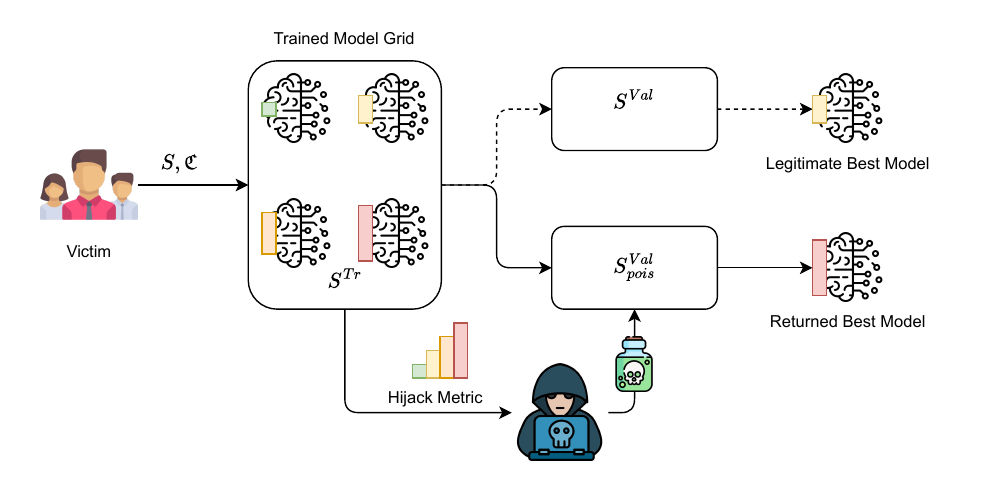}
    \caption{Schematic representation of the MOSHI threat model.}
    \label{fig:overview}
\end{figure*}

\paragraph{Adversary Knowledge.}
We now outline the adversary knowledge, detailing an attacker's information and capabilities and the limitations and constraints they face within our threat model.
We explore two distinct attack settings.
\begin{itemize}
    \item \textit{White-Box (WB) attack}: adversaries with full knowledge of the trained models $h_{\mathfrak{c}}, \;  \forall c \in \mathcal{C}$ (i.e., architectures, trained parameters and they can freely query the model for inference purposes). 
    \item \textit{Black-Box (BB) attack}: adversaries with knowledge limited to the model's architectures $\mathfrak{C}$ (i.e., the models that will be learned and tested during the model selection phase). 
\end{itemize}
Furthermore, in both scenarios, adversaries know $\mathcal{S}^{Train}$ but cannot tamper it. 
We again recall that -- for both white-box and black-box settings -- the attacker is limited to modifying the only validation set $\mathcal{S}^{Val}$.
A schematic representation of the considered threat model is depicted in Figure~\ref{fig:overview}.

\paragraph{Stealthiness.}
Poisoning only the validation set allows all the trained models and their parameters not to be modified during the attack.
This means that once attacked, the model selection phase, which is carried out in an unmodified way, returns a model that has been regularly trained on clean data, and among all the ones trained is of higher hijack metric, which the adversary can choose.
This ensures that the selected model reveals no evidence of the attack, as its performance on the victim's loss function remains close to that of models selected using the original validation data.
However, it differs regarding the hijack metric, which is influenced to align with the attacker’s desired outcome.

\subsection{Motivations and Scenarios}
\label{subsec:motivation}

\new{
The execution of the MOSHI attack can have significant repercussions for various victims by leveraging dynamics similar to traditional poisoning and backdoor attacks~\cite{tian2022comprehensive, li2022backdoor}.
One compelling scenario involves the public dissemination of open-source datasets, which are often pre-structured into standard partitions (training, validation, and testing splits) to ensure the reproducibility of experimental results.
Platforms like Hugging Face\footnote{\url{https://huggingface.co/datasets}} and Kaggle\footnote{\url{https://www.kaggle.com/}} host datasets extensively adopted by the ML community, with some datasets recording hundreds of thousands of downloads.
This popularity underscores their critical role in research and development.
However, several studies have demonstrated the feasibility of poisoning these datasets and successfully degrade model performance~\cite{carlini2024poisoning}.
Furthermore, in MLaaS, service providers often depend on client-supplied datasets, significantly increasing the risk of attackers tampering with the data.
Another scenario arises when service providers themselves act maliciously.
By providing tampered validation sets engineered using MOSHI, they could force clients to select models with higher energy consumption while maintaining optimal generalization capabilities.
This covert manipulation would increase operational costs for users while benefiting the provider financially through inflated billing, all without raising immediate suspicion.
These attacks are particularly concerning in critical applications like healthcare or autonomous systems, where inefficiencies can directly impact safety.
Moreover, in edge computing scenarios, compromised validation can result in energy-intensive models, severely affecting battery-powered devices.
By highlighting these vulnerabilities, MOSHI underscores the urgent need for robust countermeasures against adversarial interference in model selection.
}


\section{\new{Methodology}}
\label{sec:methodology}
\new{
The MOSHI attack operates uniquely by injecting and substituting data points in the validation set with data from $\mathcal{S}^{Val}_{pois}$, disrupting the critical model selection phase without altering the training process or parameters.
This set, which the attacker carefully generates, will be used for the model selection phase, which in turn will return a model $\tilde{h}_{\mathfrak{c}^*}$:
    \begin{equation}
        \label{best_poison}
 \tilde{h}_{\mathfrak{c}^*} = \argmin_{h_{\mathfrak{c}} : \mathfrak{c} \in \mathfrak{C}} \mathcal{L}_{Val}(h_{\mathfrak{c}}, \mathcal{S}^{Val}_{pois}).
    \end{equation}
The selected model $\tilde{h}_{\mathfrak{c}^*}$ is different from $h_{\mathfrak{c}^*}$, as now, the poisoned validation set no longer allows for selecting a better, more generalized, model, but selects one that has a configuration of hyper-parameters which maximizes the hijack metric, chosen by the adversary. 
Thus, a central aspect of this approach involves generating adversarial hijacking samples crafted explicitly for injection into the validation set.
Among the novelties introduced in this work, we present a specialized methodology for designing and generating these samples (Section~\ref{subsec:generation}) and the hijack metrics used in our study (Section~\ref{ssec.hm-theory}).
}



\subsection{\new{Adversarial Sample Generation}}
\label{subsec:generation}
\new{
Although our adversarial sample generation model is based on the Variational Auto Encoder (VAE) architecture (Section~\ref{subsub:vae}), we introduce a variation of the conditional VAE architecture designed for the generation of hijacking samples (Section~\ref{subsub:hvae}).
}
\subsubsection{Variational Auto Encoder (VAE)}
\label{subsub:vae}
We design our generative process using a Variational Auto Encoder (VAE)~\cite{kingma2013auto}, which is an extension of more traditional Autoencoders~\cite{hinton2006reducing}. VAE consists of two modules: first, an \textit{encoder} which learns a \textit{posterior} recognition model $q_{\phi}(z|x)$, encoding an input $x$ to a latent representation $z$; second, a \textit{decoder} that generates samples from the latent space $z$ via the likelihood model $p_{\theta}(x|z)$. $\phi$ and $\theta$ are learning parameters. 
In contrast with standard autoencoders, VAEs enforce a continuous prior distribution $p(z)$, usually set to the Gaussian. This forces the model to encode the entire input distribution to the latent code rather than memorizing single data points. 
Traditional VAEs are trained with the following loss:
\begin{equation} \small
    \mathcal{L}_{VAE}(\phi, \theta) = KL(q_{\phi}(z|x) || p(z)) 
    -\mathbb{E}_{q_{\phi}(z|x)}(\log p_{\theta}(x|z)), 
\end{equation}
where $KL$ is the Kullback-Leibler divergence~\cite{kullback1951information} that is a regularizer to keep the posterior distribution close to the prior. The second term is a simple reconstruction loss. 
For the scope of this work, we utilize a Conditional VAE (CVAE) that augments the latent space with information about the true label of a given sample~\cite{sohn2015learning}.  
\subsubsection{Hijacking VAE}
\label{subsub:hvae}
We now introduce Hijacking VAE (HVAE), a variation of the more traditional CVAE that is specifically designed to generate hijacking samples to produce $\mathcal{S}^{Val}_{pois}$.
These samples are created in such a way that, when used for computing $\mathcal{L}_{Val}$, the lower the models' hijack metric, the more significant the increase of their validation loss, hence swaying the model selection phase into returning the model that has the highest hijack metric (which has been the least penalized).
We design the HVAE loss function as follows:
    \begin{equation}
        \label{lossMHVAE}
 \mathcal{L}_{\mathrm{HVAE}} = (\mathcal{L}_{\mathrm{rec}} + \mathcal{L}_{\mathrm{KLD}} - Hj_{cost}(\mathfrak{C})) ^ 2.
    \end{equation}
Here, the terms $\mathcal{L}_{\mathrm{rec}}$ and $\mathcal{L}_{\mathrm{KLD}}$ represents the reconstruction loss and the KL divergence, as in the traditional VAE. 
The novel factor of the loss is represented by the third term $Hj_{cost}(\mathfrak{C})$.
This is the pivotal factor of the attack, defined as follows (with $\Lambda = \mathfrak{C}$):

\begin{equation} \label{cost}
     Hj_{cost}(\mathfrak{C}) = \frac{1}{|\mathfrak{C}|}\sum_{\mathfrak{c} \in \mathfrak{C}} \alpha \cdot \mathcal{L}_{Val}(h_{\mathfrak{c}}, \mathcal{S}_{gen})
\end{equation}
with 
\begin{equation} \label{alpha}
 \alpha = \frac
     {\underset{\lambda \in \Lambda}{\max} \{m(h_{\lambda}, \mathcal{S}^{Val})\} - m(h_ {\mathfrak{c}}, \mathcal{S}^{Val})}
     {\underset{\lambda \in \Lambda}{\max} \{m(h_{\lambda}, \mathcal{S}^{Val})\} - \underset{\lambda \in \Lambda}{\min} \{m(h_{\lambda}, \mathcal{S}^{Val})\}}. 
\end{equation}
 
We now explain the rationale behind Equation~\ref{cost}, which is an average of scores that are assigned to each model $\mathfrak{c} \in \mathfrak{C}$. 
The coefficient $\alpha \in \mathbb{R}$ (see Equation~\ref{alpha}) is computed by normalizing the difference between the maximum hijack metric achievable by a model $h_{\lambda}$ with $\lambda \in \Lambda = \mathfrak{C}$ and the metric of the current model.
$\alpha$ yields higher penalties the lower the hijack metric of the model $h_\mathfrak{c}$, reaching 0 if the considered model has the highest metric. This value is fixed for each model and can be computed independently of the HVAE training.
On the opposite, the second term, $\mathcal{L}_{Val}$, assesses the quality of the generative process to produce effective hijacking samples, as it computes the loss of model $h_\mathfrak{c}$ over $S_{gen}$. It is therefore computed at HVAE training time. 
\par
\begin{figure*}[!htbp] 
    \footnotesize
    \centering
    \includesvg[width=.775\textwidth]{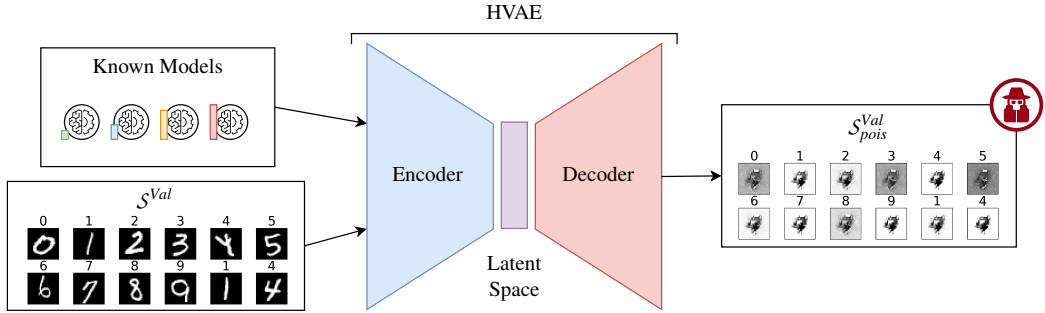}
    \caption{Schematic representation of the generation process of $\mathcal{S}^{Val}_{pois}$. For simplicity, we reported samples from the MNIST dataset~\cite{lecun2010mnist}.}
    \label{MHVAE}
\end{figure*}

Ideally, we intend to reward higher $Hj_{cost}$, as higher values imply higher losses toward those models with lower hijack metrics.
Therefore, in our loss function, we aim to maximize this value.  
During the training of the HVAE, by minimizing Equation~\ref{lossMHVAE}, we work toward:
\begin{itemize}
    \item diminishing the reconstruction loss $\mathcal{L}_{\mathrm{rec}}$, so that generated samples can resemble the original operations;
    \item diminishing the $\mathcal{L}_{\mathrm{KLD}}$ for obtaining a useful probability distribution;
    \item increasing the hijacking cost function $Hj_{cost}(\mathfrak{C})$. As the penalty value is fixed, by raising Equation~\ref{cost}, we aim at generating samples $\mathcal{S}_{gen}$, which increase the validation loss based on the magnitude of the penalty itself.
    Models with lower hijack metrics incur higher penalties, leading to increased validation loss on the generated samples. This ensures the samples are crafted to produce lower validation loss values for models with the highest hijack metrics.
\end{itemize}
A graphical overview of $\mathcal{S}^{Val}_{pois}$ generation is shown in Figure~\ref{MHVAE}.
By training the HVAE with the objective function in Equation~\ref{lossMHVAE}, the model encodes a distribution distinct from the input samples’ usual one, enabling the generation of validation samples that penalize models with lower hijack metrics.
The HVAE training procedure is detailed in Algorithm~\ref{alg.HVAE}.


\begin{algorithm}[H]
\footnotesize
    \caption{Hijack VAE Training Algorithm}
    \begin{algorithmic}[1]
        \State \textbf{Input:} HVAE model with random weights, training data $\mathcal{S}$, $\alpha_{\mathfrak{C}}$, $h_{\mathfrak{C}}$, number of epochs $epochs$
        \State \textbf{Output:} Trained HVAE model
        \For{$e \gets 1$ to $epochs$}
            \For{$\bm{x}$, $y$ in $\mathcal{S}$}  
                \State $\hat{\bm{x}} \gets $ HVAE.decode(HVAE.encode($\bm{x}$))  
                \State rec\_loss $\gets \mathcal{L}_{\mathrm{rec}}(\bm{x}, \hat{\bm{x}})$  
                \State kl\_loss $\gets \mathcal{L}_{\mathrm{KLD}}(\mathrm{HVAE})$  
                \State $\hat{\bm{x}}_{gen} \gets$ HVAE.decode(gaussian\_noise)  
                \State generated\_val\_loss $\gets \mathcal{L}_{Val}(h_{\mathfrak{C}}, \hat{\bm{x}}_{gen})$  
                \State hijack\_cost $\gets Hj_{cost}(\alpha_{\mathfrak{C}}, \mathrm{generated\_val\_loss})$  
                \State total\_loss $ \gets(\mathrm{rec\_loss + kl\_loss - hijack\_cost})^2 $  
                \State HVAE.backward\_propagation\_step(total\_loss)  
            \EndFor
        \EndFor
        \State \textbf{return} HVAE
    \end{algorithmic}
    \label{alg.HVAE}
\end{algorithm}

\subsection{Hijack Metric}
\label{ssec.hm-theory}
Generally, the purpose of a hijack metric $m$ is to produce damage to the target victim. 
\new{
We now introduce four distinct hijack metrics that impact an ML system in three different ways, i.e., generalization capabilities (Section~\ref{subsub:generalization}), latency (Section~\ref{subsub:latency}), and energy consumption (Section~\ref{subsub:energy}).
}
Note that MOSHI is not limited to such metrics, and future investigations might define different attack objectives. 

\subsubsection{\new{Generalization Capability Attack}}
\label{subsub:generalization}
This first intuitive hijack metric objective is to impact the victim model overall performance. 
Here, the objective of the attack under this metric is to choose a model that less generalizes to unseen data (e.g., test set), and therefore the result of an underfitting or overfitting training.
Therefore, this case can be considered a form of the more traditional \textit{model poisoning attack}~\cite{tian2022comprehensive}.
The metric $m$ -- that we named \textit{Generalization Metric} -- can simply compute the loss of a target model on an unseen dataset (\textit{e.g., validation set}). 
\subsubsection{Latency Attack}
\label{subsub:latency}
Increased latency in ML predictions can significantly impact the performance and usability of ML systems.
Higher latency leads to delayed responses, which can degrade user experience, particularly in real-time applications such as autonomous driving, financial trading, and interactive systems. Additionally, increased latency can hinder the efficiency of decision-making processes, as timely data processing is crucial for accurate and effective outcomes. This delay can also exacerbate the accumulation of errors, potentially compromising the reliability and accuracy of the ML model's predictions.
Therefore, an attacker might aim to induce the model selection to peak a model that results in slower predictions, on average, when deployed. 
The function $m$ -- that we named \textit{Latency Metric} -- can be designed by observing the time required by a target model to predict a set of unseen datasets (\textit{e.g., validation set}). 
\subsubsection{Energy Consumption Attack}
\label{subsub:energy}
\begin{figure}[!htbp]
    \footnotesize
    \centering
    \includesvg[width = .8\linewidth]{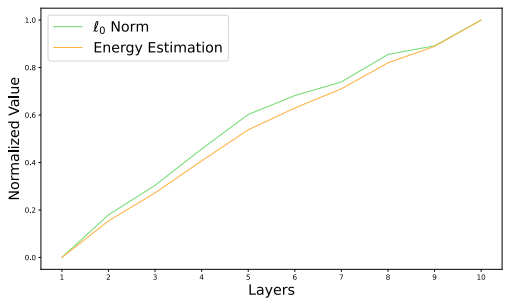}
    \caption[Histogram comparing $\ell_0$ norm and energy consumption per layer.]{Histogram comparing $\ell_0$ norm and energy consumption per layer on FFNNs from 1 to 10 layers of 32 neurons, trained on MNIST dataset with a learning rate of 0.001.}
    \label{l0_energy}
\end{figure}

Similarly to what is discussed in the motivation of the latency attack, increasing the overall energy consumption might lead to resource exhaustion. 
We inspire this metric based on the \textit{sponge attack}~\cite{shumailov2021sponge}. 
In our work, we consider two distinct metrics that measure energy consumption. 
\begin{itemize}
    \item \textit{Energy Consumption}: an estimation of the energy consumption of the model utilization that can be obtained through the OS energy consumption hosting such model. 
    \item \textit{$\ell_0$ norm}: the $\ell_0$ norm of the activations of the neurons in the network, obtained by summing the non-zero activations of each ReLU Layer in the model when it is processing a sample $\bm{x}$, then computing the mean for all samples $\bm{x} \in \mathcal{X}$. 
\end{itemize}
We opt to include this metric as \cite{cina2022energy} showed, there exists a strong link between the $\ell_0$ norm of a model and its energy consumption.
For instance, we report in Figure~\ref{l0_energy} the observed correlation between these two metrics in our experimental setting (which we will describe in the upcoming section).  

\subsection{White-box vs Black-box scenarios}
The HVAE requires knowledge about the target models, as described in Equation~\ref{cost} in the $Hj_{cost}$. Models in the grid are utilized for measuring their performance with the hijack metric and for understanding the quality of the dataset $S_{gen}$ produced by HVAE.
As we previously anticipated, in our work we consider a white-box and black-box case study. In the former, we assume the attacker has access to the exact models of the model grid. In the latter, the attacker has no such knowledge.
However, we assume that the attacker has knowledge about both training and validation sets. We can therefore leverage the \textit{adversarial transferability} of attacks. 
\par
Adversarial transferability in AML refers to the phenomenon where adversarial examples crafted to deceive one ML model can also deceive other models, even if they have different architectures or were trained on different datasets~\cite{demontis2019adversarial, alecci2023your}. This property is significant because it highlights the vulnerability of ML systems to attacks that are not specifically tailored to them, thereby posing a broader security risk.
\section{Evaluation}
\label{sec:evaluation}

\new{
In this section, we present the experimental settings of our tests (Section~\ref{subsec:experimental}) and the results of our attack (Section~\ref{subsec:results}).
}

\subsection{Experimental Settings}
\label{subsec:experimental}

\new{
We now describe the experimental settings utilized to demonstrate the MOSHI attack.
We first present the datasets and models on which we test our methodology (Section~\ref{subsub:cases}), and then we show our attack grids and HVAE implementations (Section~\ref{ssec.grid}).
Finally, we provide details on the metrics used for hijacking (Section~\ref{subsub:hmetrics}) and for the evaluation (Section~\ref{subsub:emetrics}).
}
The experiments are conducted utilizing Python 3.10 scripts, and standard ML libraries to design deep learning experiments such as PyTorch~\cite{paszke2019pytorch}. 
All experiments are executed on a machine with 32GB RAM with an NVIDIA GeForce RTX 3090.

\subsubsection{Case Studies}
\label{subsub:cases}
\begin{figure}[b]
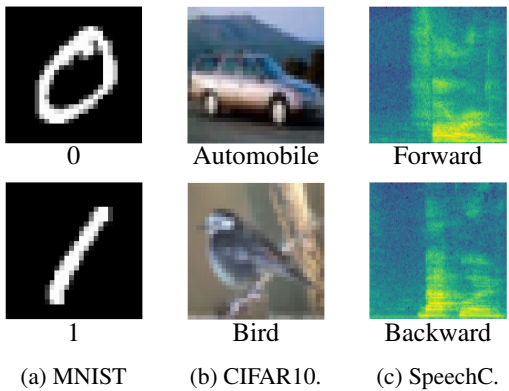

  \centering
  \begin{subfigure}{0.215\linewidth}
     \centering
     \includesvg[width=\linewidth]{figures/4-MNIST}
     \caption{MNIST}
     \label{subfig:4mnist}
  \end{subfigure}
  \hspace{0.05\linewidth}
  \begin{subfigure}{0.215\linewidth}
     \centering
     \includesvg[width=\linewidth]{figures/4-CIFAR}
     \caption{CIFAR10.}
     \label{subfig:4cifar}
  \end{subfigure}
  \hspace{0.05\linewidth}
  \begin{subfigure}{0.215\linewidth}
     \centering
     \includesvg[width=\linewidth]{figures/4-SR}
     \caption{SpeechC.}
     \label{subfig:4sr}
  \end{subfigure}
  \caption{Illustration of samples from $\mathcal{S}^{Val}$.}
  \label{fig:clean_samples}
\end{figure}

We designed three distinct case studies to test the MOSHI attack performance: two yield from the Computer Vision (CV) field, and the remaining one from Speech Recognition (SR), all in the classification task.
We opted to test these two distinct application domains to understand the generalization of the attack in different AI scenarios. 
We now briefly describe the three tasks.\footnote{All datasets were imported using the corresponding PyTorch libraries.} 
\begin{itemize}
    \item \textit{MNIST (CV)~\cite{lecun2010mnist}}. It is a dataset of bilevel $28 \times 28$ handwritten digits divided into 10 classes: one for each digit from 0 to 9. The victim models considered for this case are FeedForward Neural Networks, with hyperparameters grid concerning the number of neurons per layer and the number of layers themselves. 
    \item \textit{CIFAR10 (CV)~\cite{cifar10}}. It is a dataset of colored $32 \times 32$ images of various subjects belonging to 10 classes: airplane, automobile, bird, cat, deer, dog, frog, horse, ship, and truck. In this case study the victim models are DenseNets \cite{huang2017densely}, of which we adapted the implementation done by \cite{PyTorchDenseNet}. The considered hyperparameters grid sees the number of dense blocks and the number of neurons for the fully connected layer which feeds the output one.
    \item \textit{Speech Commands (SR)~\cite{speechcommandsdataset}}. It is a dataset composed of audio files that are about 1 second long (around 16000 frames long) and belong to 35 classes of different commands spoken by different people. Here, the victim models are Convolutional Neural Networks composed of 1D convolutional layers, and the considered hyperparameters are the number of convolutional layers and the width of the fully connected layer that feeds the output one (code adapted from \cite{PyTorchSpeechCommands}).
\end{itemize}
All the models considered are feedforward neural networks, with an output composed of one or more dense layers that map the hidden representations to the output space.
The architecture of the hidden layers varies depending on the specific task. A detailed list of the layers used is provided in Table \ref{tab.hp}.
In total, we trained 472 models, considering both white-box and black-box scenarios, 360 for MNIST, 84 for CIFAR10, and 32 for Speech Commands. 
Figure~\ref{fig:clean_samples} shows two examples for each dataset. Note that for the Speech Commands, we display the frequency spectrum. 

\begin{table}[!htpb]
\centering
\footnotesize
\caption{Hyper parameters grouped by use case. }
\begin{tabular}{cc} \toprule
\multicolumn{2}{c}{\textit{\textbf{MNIST}}} \\ \midrule
    $\#$ layers &     $[1, 2, 3, 4, 5, 6, 7, 8, 9, 10]$        \\
    $\#$ neurons &     $[32, 64, 128]$        \\
    $\#$ learning rate &     $[0.0001, 0.0005, 0.001, 0.005, 0.01, 0.05]$       \\ \midrule
    \multicolumn{2}{c}{\textit{\textbf{CIFAR10}}} \\ \midrule
    $\#$ dense blocks &     $[2, 3, 4, 5, 6, 7, 8]$        \\
    $\#$ neurons dense layer&     $[128, 256, 512]$        \\
    $\#$ learning rate &     $[0.001, 0.005]$       \\ \midrule
    \multicolumn{2}{c}{\textit{\textbf{Speech Commands}}} \\ \midrule
    $\#$ 1D conv layers &     $[2, 4, 6, 8]$        \\
    $\#$ neurons dense layer &     $[128, 256]$        \\
    $\#$ learning rate &     $[0.001, 0.005]$       \\ \bottomrule
\end{tabular}
\label{tab.hp}
\end{table}


\subsubsection{Attack Grid}
\label{ssec.grid}
Regardless of the case study at hand, we conduct the attacks by considering each task's whole model's grids but also subsets. 
For instance, in the MNIST case, we execute the attack when considering the whole hyperparameters grid (i.e., the number of layers, number of neurons, learning rate) but also in cases where we fixed - for instance - the learning rate and we vary the other two hyperparameters. 
With this experimental setting, we aim to understand the attack effectiveness at different granularity.
Table~\ref{tab.grid} describes the attack grid. 
For instance, when considering the MNIST scenario, we can execute the attack on the entire combination of hyperparameters (\textit{i.e.} learning rate, number of layers and neurons), consisting in one set containing 180 models, or, when grouping on the learning rate, we obtain 6 distinct sets (one per learning rate) constituted by 30 models ($3$ types of neurons $\times$ $10$ types of number of layers). 

\begin{table}[!htpb]
\centering
\footnotesize
\caption{Attack grid grouped by use case. }
\begin{tabular}{ccc} \toprule
\multicolumn{3}{c}{\textit{\textbf{MNIST}}} \\ \midrule
    \textit{Grouped-by} & \textit{\# sets} & \textit{\# models}\\ \hline
    Global &     1    &  180 \\
    learning rate &     6  &  30   \\
    learning rate \& $\#$ neurons &  18 &  10\\ \midrule
    \multicolumn{3}{c}{\textit{\textbf{CIFAR10}}} \\ \midrule
    Global &  1  &  42 \\
    learning rate &  2   & 21\\
    learning rate \& $\#$ neurons &    6    &   7 \\ \midrule
    \multicolumn{3}{c}{\textit{\textbf{Speech Commands}}} \\ \midrule
    Global &     1   &  16\\
    learning rate &   2  &  8 \\ 
    learning rate \& $\#$ neurons &    4   &  4  \\ \bottomrule
\end{tabular}
\label{tab.grid}
\end{table}
\par
In both white-box and black-box settings, we assume that the attacker has some different amount of knowledge at his disposal when launching the attack.
In other words, we used sets of the models, either clean or surrogate (obtained the same way as the aforementioned attack grid), to train the HVAE.
In total, we trained 50 HVAE in MNIST: 25 sets over the three grouping conditions multiplied by the two knowledge levels i.e., white-box and black-box).
Similarly, we trained 18 and 14 HVAE for CIFAR10 and Speech Commands, respectively. 
\par
Finally, for each dataset, we trained HVAE with different architectures: 
\begin{itemize}
    \item \textit{MNIST}. The HVAE comprises only fully connected layers and is trained using Binary Cross Entropy as a reconstruction loss ($\mathcal{L}_{\mathrm{rec}}$). 
    \item \textit{CIFAR10}. For CIFAR10, the HVAE adopted is composed of 2D convolutional layers, and the adopted reconstruction loss ($\mathcal{L}_{\mathrm{rec}}$) is the Mean Square Error.
    \item \textit{Speech Commands}. The HVAE are made of 1D convolutional layers, with the Mean Square Error as the reconstruction loss ($\mathcal{L}_{\mathrm{rec}}$).
\end{itemize}

\subsubsection{Hijack Metrics}
\label{subsub:hmetrics}
We now describe the implementation details of the hijack metrics described in Section~\ref{ssec.hm-theory}. 
\begin{itemize}
    \item \textit{Generalization Metric}. This hijack metric consists of the loss obtained by our models when processing the original (legitimate) validation set. 
    \item \textit{Latency Metric}. It is obtained by measuring the overall processing time of a given set of input data.
    \item \textit{Energy Consumption}. it is computed by an ASIC simulator \footnote{\url{https://github.com/iliaishacked/sponge_examples}} \footnote{\url{https://github.com/Cinofix/sponge_poisoning_energy_latency_attack}} adapting the functions already developed for \cite{shumailov2021sponge} and \cite{cina2022energy}. As this tool was not compatible with the specific architecture used for the classification models of the SpeechCommands dataset, this was the only case in which we did not consider this specific metric.
    \item \textit{$\ell_0$ Norm}. we computed the mean of the non-zero activations of each ReLU Layer in our models when processing a set of input data.
This metric is closely related to energy consumption and can be considered an alternative that does not rely on ASIC simulators. As a result, it can be applied to any architecture, offering broader applicability.
\end{itemize}

\subsubsection{Evaluation Metrics}
\label{subsub:emetrics}
In order to qualitatively estimate the effectiveness of the
attack against the specific model’s grid set considered, we
introduce two novel distinct metrics: the \textit{Effectiveness Score Function} (ESF) and the \textit{Attack Success Rate} (ASR).
\par
The ESF goal is to understand the impact of MOSHI on the victim model when deployed.
Therefore, with ESF we measure the attack returned model $\tilde{h}_{\mathfrak{c}^*}$ and the legitimate $h_{\mathfrak{c}^*}$ based on the hijack metric $\mathcal{E}$.
Formally, ESF is defined as follows:
\begin{equation}
     \mathscr{S}(\mathfrak{C}) = \frac{
         \mathcal{E} ( \tilde{h}_{\mathfrak{c}^*}, \mathcal{S}^{Test} ) 
         - 
         \mathcal{E} ( h_{\mathfrak{c}^*}, \mathcal{S}^{Test} )
     }
     {
         \underset{\mathfrak{c} \in \mathfrak{C}}{\max} \{\mathcal{E}(h_{\mathfrak{c}}, \mathcal{S}^{Test})\} - 
         \mathcal{E} (    h_{\mathfrak{c}^*}, \mathcal{S}^{Test}    )
     }.
\end{equation}

ESF takes the difference between a chosen metric $\mathcal{E}$ of the models $\tilde{h}_{\mathfrak{c}^*}$ (see Equation~\ref{best_poison}) and $h_{\mathfrak{c}^*}$ (see Equation~\ref{best_val}) normalizing it with the difference between the maximum possible chosen metric of a model $h_\mathfrak{c}$ with $\mathfrak{c} \in \mathfrak{C}$, and again the chosen metric by $h_{\mathfrak{c}^*}$.\\
This function allows us to evaluate our attack with any metric $\mathcal{E}$, for the purpose of this study, with regards to the hijack metric considered:
\begin{itemize}
    \item \textit{Generalization Metric}. $\mathcal{E}$ is the loss function computed on the original validation set;
    \item \textit{Latency Metric}. $\mathcal{E}$ is the latency while processing the test set.
    \item \textit{Energy Consumption} or \textit{$\ell_0$ Norm}. $\mathcal{E}$ is the estimated energy consumed by the model while processing the test set. When testing the attack in the SpeechCommands case, as we could not compute the \texttt{Energy} metric, we evaluated the quality of the attack using the \texttt{$\ell_0$ Norm}, by adopting $\mathcal{E}$ to be the $\ell_0$ Norm.
\end{itemize}
This Effectiveness Score function allows us to give a normalized score that sums up the effectiveness of the our attack, as a matter of fact:
\begin{itemize}
    \item \textit{Successful}, when $ESF > 0$, the MS phase to return a model with higher $\mathcal{E}$ compared to $h_{\mathfrak{c}^*}$. If the score is equal to $1.0$, then the returned model $\tilde{h}_{\mathfrak{c}^*}$ is the one among the considered models to have the highest $\mathcal{E}$.
    \item \textit{Invariant}, when $ESF = 0.0$, the attack did not sway the MS phase, and the returned model is $\tilde{h}_{\mathfrak{c}^*} = h_{\mathfrak{c}^*}$, \textit{i.e.} the returned model is also the legitimate one. 
    \item \textit{Unsuccessful}, $ESF < 0.0$ imply that the attack made the MS phase return a model of less $\mathcal{E}$ compared to $h_{\mathfrak{c}^*}$.
\end{itemize}  
\par
We define the ASR as a simplification of the ESF. In particular, we define the ASR as a boolean variable of a successful or unsuccessful attack, and therefore the $ASR = 1$ if $ESF > 0$, $ASR = 0$ otherwise. 
With this metric, we are interested in understanding when an attack is successful in terms of increasing the hijack metric.

\subsection{Results}
\label{subsec:results}

We analyze the impact of the MOSHI attack on the experimental settings previously described.
As our attack generates malicious poisoning samples to inject into the validation set, we explore the attack effect in two fashions.
\begin{itemize}
    \item \textit{Full substitution}, where the validation set is entirely substituted by the malicious samples (Section~\ref{subsub:full}). 
    \item \textit{Partial substitution}, where the validation set combines both legitimate and malicious samples at different rates, i.e. 10\%, 20\%, 50\%, 80\%, 100\% (Section~\ref{subsub:partial}). 
\end{itemize}
In Section~\ref{subsub:impact} we measure the attack performance by comparing -- for each grid attack presented in Table~\ref{tab.grid} -- the legitimate model (i.e., the model with the lowest loss in the unpoisoned validation set) and the malicious model (i.e., the model with the lowest loss in the poisoned validation set). 

\subsubsection{Full Substitution}
\label{subsub:full}

Table~\ref{tab.asr-wb} and Table~\ref{tab.asr-bb} show the results of MOSHI in both white-box and black-box, respectively.
Each table is composed of three sub-tables that represent the three distinct attack grid granularities, i.e., global, grouped-by learning rate, and grouped-by learning rate and number of neurons.
For each case, we report the results for the three case studies (i.e., MNIST, CIFAR10, and Speech Commands) spanning across the four studied hijack metrics (i.e., generalization error, latency, energy, and $\ell_0$). 
Furthermore, each row reports the number of sets.\footnote{Each set we trained an ad-hoc HVAE (see Section~\ref{ssec.grid}).}
Consider Table~\ref{tab.asr-bb} on the MNIST use-case and global granularity: since we have only 1 set, the attack is either successful (100\%) or unsuccessful (0\%).
\par
Overall, we denote strong positive results proving the validity of HVAE. 
The attack is perfectly successful (i.e., $ASR = 100\%$) in $22/33$ and $23/33$ cases in both white-box and black-box scenarios.
Interestingly, we do not observe differences between white and black box settings, implying that the attack can be very dangerous, yielding high transferability capabilities. 
\new{
HVAE also proves particularly damaging to generalization: in every evaluated instance, the poisoned validation set successfully hijacked the model selection process, consistently leading to the selection of suboptimal models with inferior generalization performance.
This highlights the implications of the proposed attack MOSHI on real-world ML pipelines.
}
\par
Last, while our proposed attack HVAE shows promising results across different benchmarks, it suffers from the increased complexity of the data.
In particular, in the SpeechCommands, the attack fails to produce poisoned validation samples capable of hijacking the model selection.   
This phenomenon may be attributed to the underlying architecture of the HVAE, which is not well-suited for learning effective representations in sequential domains.
We believe that future studies might attempt to design ad hoc HVAE for specific tasks (e.g., RAVE for the speech domain~\cite{caillon2021rave}).
\par
We can conclude that our proposed attack is effective and can produce strong manipulations to the model selection phase. 
Furthermore, these results answer our original research question: \textit{can an attacker manipulate the model selection?}
The answer is yes, we can produce a MOSHI attack, and an attacker can ideally choose custom hijack metrics.

\begin{table}[!htpb]
\centering
\footnotesize
\caption{ASR in White-Box settings. at different attack grid granularities.}
\begin{tabular}{c|c|c|c|c|c}  \toprule
\multicolumn{6}{c}{\textit{\textbf{Global}}} \\ \midrule
 & \textit{\# Sets} &\textit{Gener.} & \textit{Latency} & \textit{Energy} & \textit{$\ell_0$} \\ \midrule
MNIST & 1 & 100.0\% & 100.0\% & 100.0\% & 100.0\% \\
CIFAR10 & 1 &100.0\% & 0.0\% & 100.0\% & 100.0\% \\
SpeechC. & 1 & 100.0\% & 0.0\% & N/A & 0.0\% \\ \midrule
\multicolumn{6}{c}{\textit{\textbf{Learning Rate}}} \\ \midrule
MNIST & 6 &  100.0\% & 100.0\% & 66.7\% & 83.3\% \\
CIFAR10 & 2 & 100.0\% & 100.0\% & 100.0\% & 100.0\% \\
SpeechC. & 2 & 100.0\% & 50.0\% & N/A & 0.0\% \\\midrule
\multicolumn{6}{c}{\textit{\textbf{Learning Rate \& \# Neurons}}} \\ \midrule
MNIST & 18 & 100.0\% & 94.4\% & 100.0\% & 100.0\% \\
CIFAR10 & 6 & 100.0\% & 33.3\% & 50.0\% & 50.0\% \\
SpeechC. & 4 & 100.0\% & 50.0\% & N/A & 0.0\% \\ \bottomrule
\end{tabular}
\label{tab.asr-wb}
\end{table}

\begin{table}[!htpb]
\centering
\footnotesize
\caption{ASR in Black-Box Settings at different attack grid granularities.}
\begin{tabular}{c|c|c|c|c|c}  \toprule
\multicolumn{6}{c}{\textit{\textbf{Global}}} \\ \midrule
 & \textit{\# Sets} & \textit{Gener.} & \textit{Latency} & \textit{Energy} & \textit{$\ell_0$} \\ \midrule
MNIST & 1 & 100.0\% & 100.0\% & 100.0\% & 100.0\% \\
CIFAR10 & 1   & 100.0\% & 100.0\% & 100.0\% & 100.0\% \\
SpeechC.  & 1 & 100.0\% & 0.0\% & N/A & 0.0\% \\ \midrule
\multicolumn{6}{c}{\textit{\textbf{Learning Rate}}} \\ \midrule
MNIST & 6 &  100.0\% & 100.0\% & 66.7\% & 83.3\% \\
CIFAR10 & 2 & 100.0\% & 50.0\% & 50.0\% & 100.0\% \\
SpeechC. & 2 & 100.0\% & 50.0\% & N/A & 0.0\% \\ \midrule
\multicolumn{6}{c}{\textit{\textbf{Learning Rate \& \# Neurons}}} \\ \midrule
MNIST & 18 & 100.0\% & 83.3\% & 100.0\% & 100.0\% \\
CIFAR10 & 6 & 83.3\% & 50.0\% & 33.3\% & 50.0\% \\
SpeechC. & 4 & 100.0\% & 50.0\% & N/A & 50.0\% \\ \bottomrule
\end{tabular}
\label{tab.asr-bb}
\end{table}

\subsubsection{Partial Substitution}
\label{subsub:partial}

\begin{figure*}[!htpb]
  \centering
  \begin{subfigure}{0.495\linewidth}
    \centering
    \includegraphics[width=\linewidth]{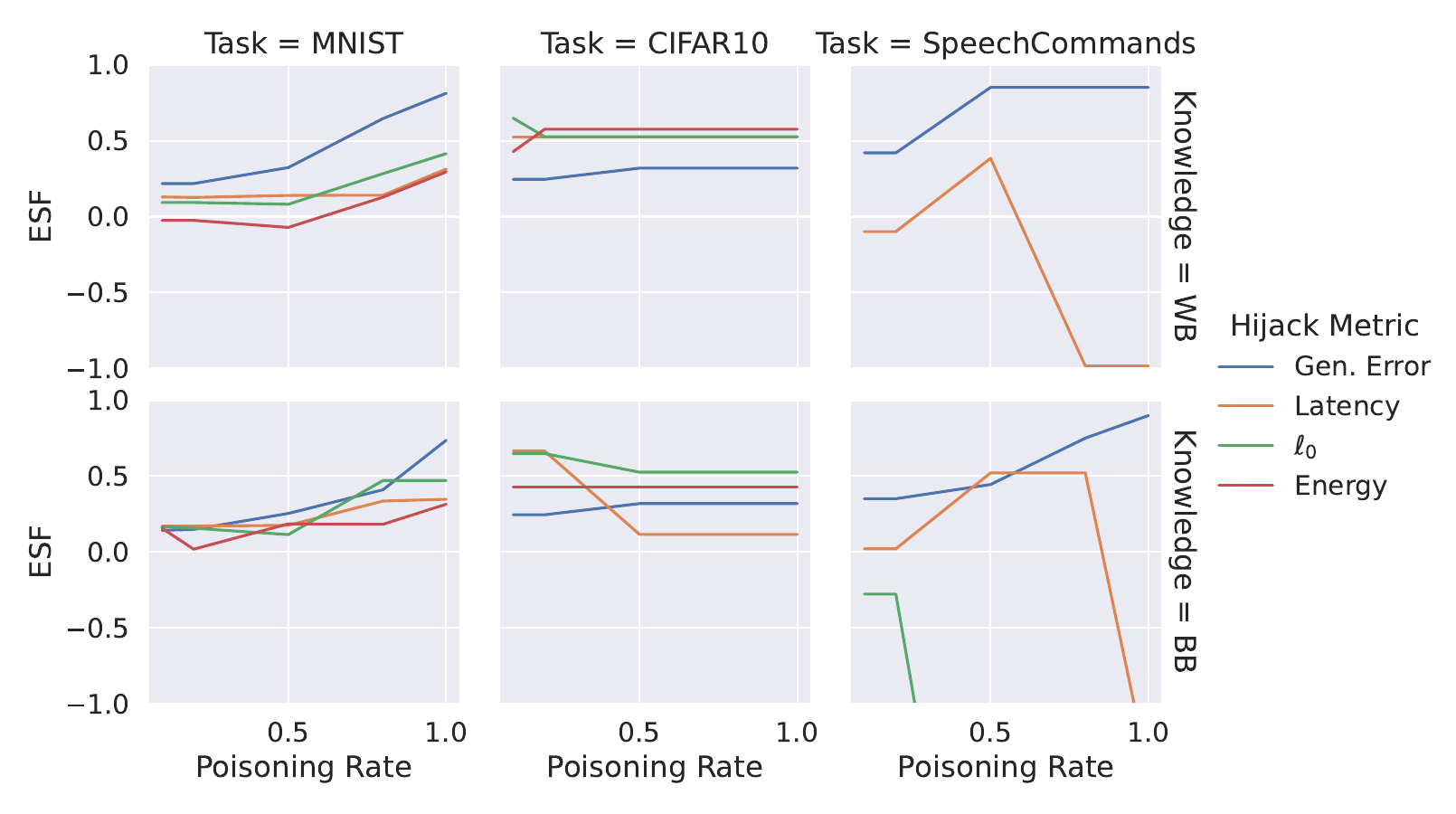}
    \caption{``\textit{Learning rate}'' attack grid setting.}
    \label{fig:pr-impact-lr}
  \end{subfigure}
  \begin{subfigure}{0.495\linewidth}
    \centering
    \includegraphics[width=\linewidth]{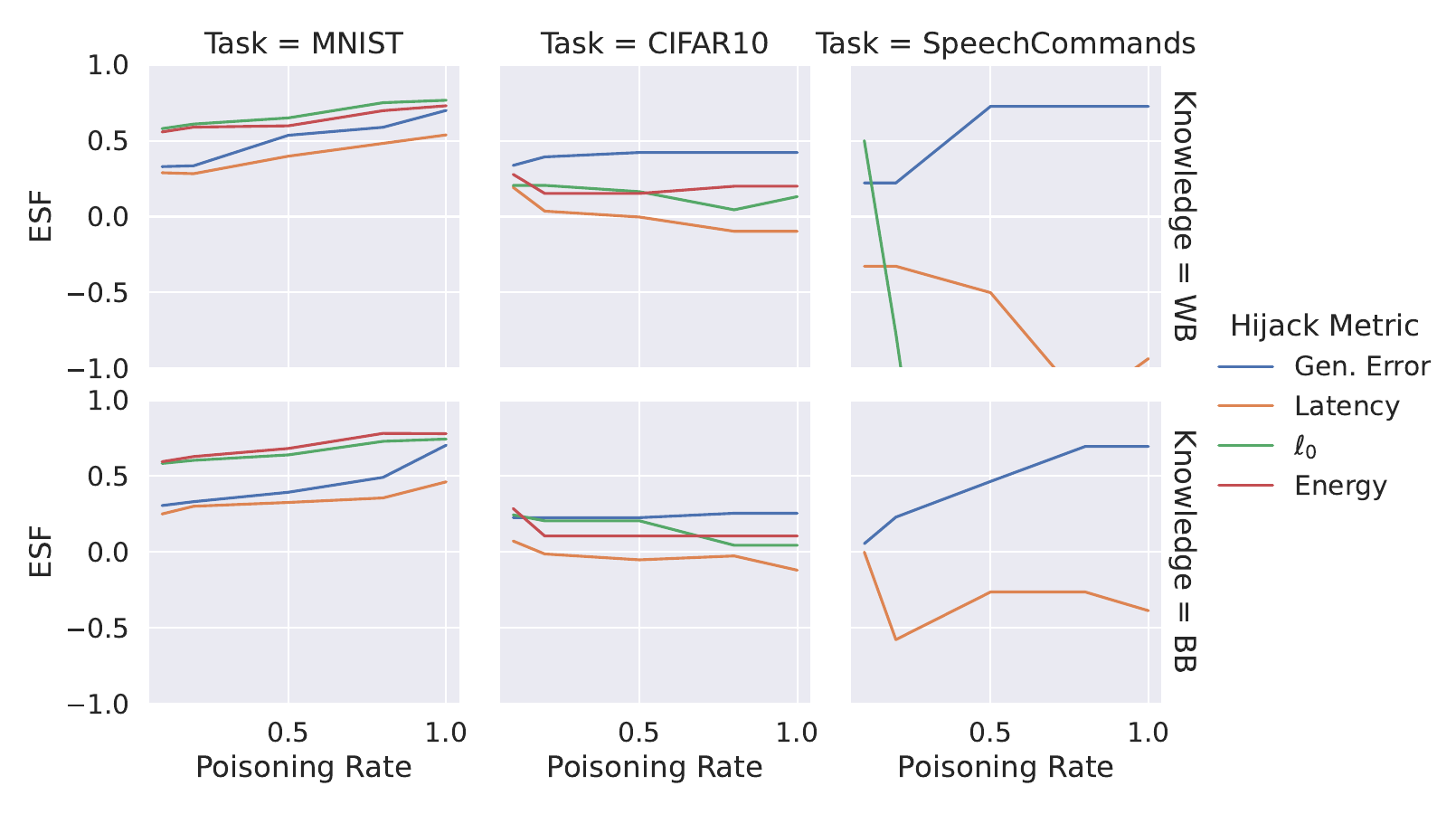}
    \caption{``\textit{Learning rate \& \#neurons}'' attack grid setting.}
    \label{fig:pr-impact}
  \end{subfigure}
  \caption{White-Box (WB) and Black-Box (BB) poisoning rate impact.}
  \label{fig:impact}
\end{figure*}

We now analyze the impact on the validation poisoning rate. 
Figure~\ref{fig:impact} shows such a trend on the grid attack settings \textit{learning rate} and \textit{learning rate \& \#neurons}, respectively, where we report the three use cases, the four hijack metrics, and both knowledge levels.
By analyzing the ESF for the first two datasets considered, we can observe that the white-box settings help produce a more potent attack comparable to more traditional poisoning properties.
Despite that, we do not observe clear advantages in poisoning the validation set entirely. This result suggests that HVAE can effectively attack even by tampering with only a smaller portion of the validation set.
On the other hand, in the Speech Commands case study, we observe a more erratic behavior of the mean ESF.
This is particularly evident for the $\ell_0$ hijack metric, despite achieving an ASR greater than zero, as illustrated in Figure~\ref{fig:pr-impact-lr} and Figure~\ref{fig:pr-impact}, the mean ESF of these attacks is heavily below 0.
This discrepancy arises because, for some attacks, the ESF values are strongly negative.
We believe that the interplay between MOSHI effectiveness and the poisoning rate should be explored in future works. 

 


\subsubsection{\new{Attack Impact}}
\label{subsub:impact}
In this section, we present the impact of our attack on the targeted metrics for the specific case study and models considered. 
In Table~\ref{tab.impact}, we illustrate how MOSHI, by influencing the Model Selection phase, impacts the chosen metrics.
This is done by reporting the ratio between the metrics of the model returned by an untampered Model Selection, and the model returned once the attack, conducted by training the HVAE with full knowledge of the victim models and with full (100\%) substitution of the original validation set with poison data. 
For instance, when considering the white-box knowledge, we observed a malicious MNIST model $\times3$ slower than the legitimate model in terms of latency introduced.
As we can see, the WB and BB settings perform similarly in both the MNIST and SpeechCommands case studies.
The former presents the same model for all the attacks conducted with the various hijack metrics, and the latter shows better performance only for the WB case on the Generalization Metric.
Finally, for the CIFAR10 case study, the BB setting performs slightly better, allowing for a small increase in the Latency metric.

\begin{table}[!htpb]
\centering
\footnotesize
\caption{Impact factor on the availability. The results report the impact factor. We only report the values for the \textit{global} attacker grid variant. }
\begin{tabular}{c|c|c|c|c}  \toprule
\multicolumn{5}{c}{\textit{\textbf{White-Box}}} \\ \midrule
 &\textit{Gener.} & \textit{Latency} & \textit{Energy} & \textit{$\ell_0$} \\ \midrule
MNIST  & 21.27 & 3.818 & 1.244 & 6.293 \\
CIFAR10 & 1.120 & 0.769 & 1.009 & 1.027 \\
SpeechC. & 4.136 & 0.797 & N/A & 0.629 \\ \midrule
\multicolumn{5}{c}{\textit{\textbf{Black-box}}} \\ \midrule
MNIST  & 21.27 & 3.818 & 1.244 & 6.293 \\
CIFAR10 & 1.180 & 1.001 & 1.009 & 1.027 \\
SpeechC. & 2.319 & 0.797 & N/A & 0.810 \\ \bottomrule
\end{tabular}
\label{tab.impact}
\end{table}

\new{
Evaluating our attack's impact on availability reaffirms the strong correlation between the attack's effectiveness and the diversity of models involved in the selection process.
A more extensive and diverse set of models enhances the potential for achieving higher performance metrics, enabling the training of more effective HVAE models and, consequently, executing a more impactful attack.
}
\section{\new{Discussion}}
\label{sec:discussion}
\new{
We now analyze HVAE behavior in terms of the produced adversarial sample (Section~\ref{subsec:quality}), and loss function (Section~\ref{subsec:objective}).
We also discuss possible countermeasures for our MOSHI attack (Section~\ref{subsec:counter}).
}

\subsection{Qualitative Analysis}
\label{subsec:quality}
\begin{figure}[!htpb]
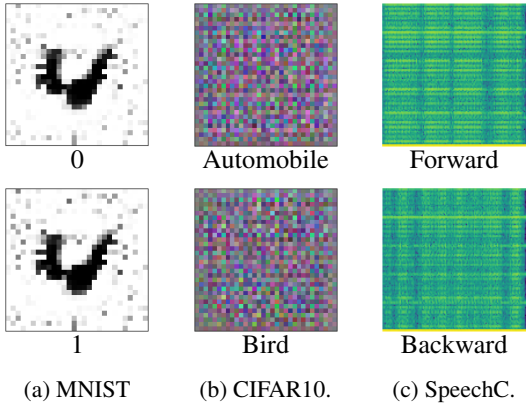

  \centering
  \begin{subfigure}{0.225\linewidth}
     \centering
     \includesvg[width=\linewidth]{figures/7-MNIST.svg}
     \caption{MNIST}
     \label{subfig:7mnist}
  \end{subfigure}
  \hspace{0.05\linewidth}
  \begin{subfigure}{0.225\linewidth}
     \centering
     \includesvg[width=\linewidth]{figures/7-CIFAR.svg}
     \caption{CIFAR10.}
     \label{subfig:7cifar}
  \end{subfigure}
  \hspace{0.05\linewidth}
  \begin{subfigure}{0.225\linewidth}
     \centering
     \includesvg[width=\linewidth]{figures/7-SR.svg}
     \caption{SpeechC.}
     \label{subfig:7sr}
  \end{subfigure}
  \caption{Illustration of samples from $\mathcal{S}^{Val}_{pois}$.}
  \label{fig:poison_samples}
\end{figure}

Here we will discuss the quality samples generated by our HVAE, which constitute the poisoned validation set $\mathcal{S}^{Val}_{pois}$.
\new{
As an example, Figure~\ref{fig:poison_samples} showcases two samples for each dataset, along with their corresponding labels, generated by the HVAE trained on the MNIST, CIFAR10, and SpeechC cases.
}
As it can be easily seen, these poisoned samples carry little to no resemblance to the original samples.
Such a result suggests that HVAE objective loss (see Equation~\ref{lossMHVAE}) does take little consideration reconstruction loss $\mathcal{L}_{\mathrm{rec}}$. 
We investigate the interplay between the three components (\textit{i.e., $\mathcal{L}_{\mathrm{rec}}, \mathcal{L}_{\mathrm{KLD}}, Hj_{cost}$}) in Section~\ref{subsec:objective}.


\subsection{Objective Function}
\label{subsec:objective}

\begin{figure}[!htbp]
    \centering
    \includesvg[width=0.8\linewidth]{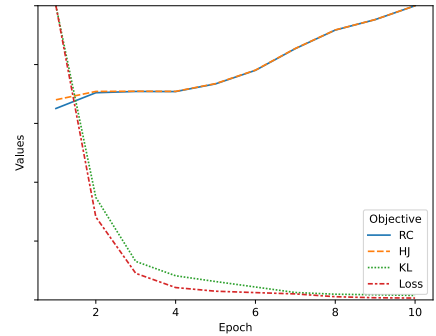}
    \caption{HVAE objective function for: MNIST dataset, White-box knowledge, $\ell_0$, and global granularity grid. Note that the image is meant to show only the objective trends and it does not reflect the real values.}
    \label{HVAE_losses}
\end{figure}

In this section, we investigate the interplay of the factors that compose the objective described in Equation~\ref{MHVAE}, on different tasks and different hijack metrics. 
This mainly happens as the two factors $\mathcal{L}_{\mathrm{rec}}$ and $Hj_{cost}$ work one against the other.
In order to better visualize this phenomenon, we report Figure~\ref{HVAE_losses}, depicting the trend of both factors during the training of an HVAE, assuming the complete knowledge of the 180 models trained for the MNIST case.
From this figure, it can be seen how both factors follow
the same trend.
Therefore, aiming at increasing the $Hj_{cost}$ for creating poisonous samples able to sway the MS phase is, at the same time, contrasting the generation of samples that resemble the original ones.

\subsection{\new{Countermeasures}}
\label{subsec:counter}
MOSHI-generated samples do not resemble the original training points qualitatively speaking. 
This represents a weakness in MOSHI, which we hypothesize could allow a defense, that focuses on the detection of the poisonous samples - like MagNet~\cite{2017magnet} - to be able to contrast it.
Usually, defenses of this type are implemented against evasion attacks and are able to detect or perform small corrections on input samples, before they are presented to the network.
To our knowledge, no defense mechanism of this kind has ever been deployed to defend the Model Selection phase (as MOSHI is the first of its kind); moreover, we hypothesize that such a defense may be able to contrast the attack as is, but, we also believe it possible, that future studies may be able to create samples better resembling the clean ones, thus rendering these defenses ineffective.
\section{Conclusion}
\label{sec:conclusion}

\new{
Model selection is a cornerstone of ML, directly shaping a system's performance, robustness, and generalizability.
Selecting the right model ensures accurate data interpretation, reliable predictions, and adaptability to new scenarios.
Additionally, this paper we demonstrates how model selection can be used to negatively impact the efficiency and scalability of real-world applications, making careful evaluation a critical step in the ML pipeline.
}

\new{
This work addresses a novel question in AML: \textit{can an attacker tamper with the validation set to manipulate model selection arbitrarily?}
To answer this, we introduce the MOSHI (MOdel Selection HIjacking) adversarial attack, the first of its kind, which targets the model selection phase by injecting adversarial samples into the validation set.
This manipulation prioritizes the hijack metric while leaving the victim’s original loss metric and code untouched.
To achieve this goal, we introduce an innovative generative process called the \textit{Hijaking VAE} (HVAE), designed to craft adversarial samples that enable the attack.
Our comprehensive experiments validate the feasibility of arbitrarily influencing model selection, highlighting the broad applicability of MOSHI due to its reliance solely on data injection, independent of the victim’s model or validation methodology.
}

\section*{Ethics Considerations}
In this work, we propose a novel adversarial machine learning attack and conduct experiments exclusively in controlled, lab-only environments. Our primary goal is to advance the understanding of adversarial vulnerabilities in machine learning systems and to contribute to the development of more robust defenses. All experiments were carried out with benchmark datasets or models designed for research purposes, ensuring no real-world harm or exploitation of sensitive data. We acknowledge the dual-use nature of adversarial research and emphasize our commitment to ethical guidelines by openly discussing mitigation strategies and encouraging responsible use of this work.
\section*{Open Science}

All the resources required for reproducing the experiments described in this study are provided.
This includes the complete dataset (traditional benchmarks), pre-processing scripts, model training code, and detailed instructions for setting up the computational environment. By ensuring accessibility to these resources, we aim to facilitate transparency, reproducibility, and further research in the field of ML.
All supplementary materials are available in the accompanying repository, providing a comprehensive framework for replicating and validating our findings.

\balance
\bibliographystyle{plain}
\bibliography{bibliography}

\begin{thebibliography}{10}

\bibitem{PyTorchSpeechCommands}
{\em \url{https://pytorch.org/}}, 2024-10-10.
\newblock \\URL: {\url{https://pytorch.org/tutorials/intermediate/speech_command_recognition_with_torchaudio.html}}.

\bibitem{alecci2023your}
Marco Alecci, Mauro Conti, Francesco Marchiori, Luca Martinelli, and Luca Pajola.
\newblock Your attack is too dumb: Formalizing attacker scenarios for adversarial transferability.
\newblock In {\em Proceedings of the 26th international symposium on research in attacks, intrusions and defenses}, pages 315--329, 2023.

\bibitem{barreno2006can}
Marco Barreno, Blaine Nelson, Russell Sears, Anthony~D. Joseph, and J.~D. Tygar.
\newblock Can machine learning be secure?
\newblock In {\em Proceedings of the 2006 ACM Symposium on Information, Computer and Communications Security}, ASIACCS '06, page 16–25, New York, NY, USA, 2006. Association for Computing Machinery.

\bibitem{biggio2013evasion}
Battista Biggio, Igino Corona, Davide Maiorca, Blaine Nelson, Nedim {\v{S}}rndi{\'{c}}, Pavel Laskov, Giorgio Giacinto, and Fabio Roli.
\newblock Evasion attacks against machine learning at test time.
\newblock In Hendrik Blockeel, Kristian Kersting, Siegfried Nijssen, and Filip {\v{Z}}elezn{\'y}, editors, {\em Machine Learning and Knowledge Discovery in Databases}, pages 387--402, Berlin, Heidelberg, 2013. Springer Berlin Heidelberg.

\bibitem{biggio2012poisoning}
Battista Biggio, Blaine Nelson, and Pavel Laskov.
\newblock Poisoning attacks against support vector machines.
\newblock In {\em Proceedings of the 29th International Coference on International Conference on Machine Learning}, ICML'12, page 1467–1474, Madison, WI, USA, 2012. Omnipress.

\bibitem{caillon2021rave}
Antoine Caillon and Philippe Esling.
\newblock Rave: A variational autoencoder for fast and high-quality neural audio synthesis.
\newblock {\em arXiv preprint arXiv:2111.05011}, 2021.

\bibitem{carlini2024poisoning}
Nicholas Carlini, Matthew Jagielski, Christopher~A Choquette-Choo, Daniel Paleka, Will Pearce, Hyrum Anderson, Andreas Terzis, Kurt Thomas, and Florian Tram{\`e}r.
\newblock Poisoning web-scale training datasets is practical.
\newblock In {\em 2024 IEEE Symposium on Security and Privacy (SP)}, pages 407--425. IEEE, 2024.

\bibitem{chung2018serving}
Eric Chung, Jeremy Fowers, Kalin Ovtcharov, Michael Papamichael, Adrian Caulfield, Todd Massengill, Ming Liu, Daniel Lo, Shlomi Alkalay, Michael Haselman, et~al.
\newblock Serving dnns in real time at datacenter scale with project brainwave.
\newblock {\em iEEE Micro}, 38(2):8--20, 2018.

\bibitem{cina2022energy}
Antonio~Emanuele Cin{\`a}, Ambra Demontis, Battista Biggio, Fabio Roli, and Marcello Pelillo.
\newblock Energy-latency attacks via sponge poisoning.
\newblock {\em arXiv preprint arXiv:2203.08147}, 2022.

\bibitem{demontis2019adversarial}
Ambra Demontis, Marco Melis, Maura Pintor, Matthew Jagielski, Battista Biggio, Alina Oprea, Cristina Nita-Rotaru, and Fabio Roli.
\newblock Why do adversarial attacks transfer? explaining transferability of evasion and poisoning attacks.
\newblock In {\em 28th USENIX security symposium (USENIX security 19)}, pages 321--338, 2019.

\bibitem{goodfellow2016deep}
Ian Goodfellow, Yoshua Bengio, and Aaron Courville.
\newblock {\em Deep Learning}.
\newblock MIT Press, 2016.

\bibitem{hazelwood2018applied}
Kim Hazelwood, Sarah Bird, David Brooks, Soumith Chintala, Utku Diril, Dmytro Dzhulgakov, Mohamed Fawzy, Bill Jia, Yangqing Jia, Aditya Kalro, et~al.
\newblock Applied machine learning at facebook: A datacenter infrastructure perspective.
\newblock In {\em 2018 IEEE international symposium on high performance computer architecture (HPCA)}, pages 620--629. IEEE, 2018.

\bibitem{hinton2006reducing}
Geoffrey~E Hinton and Ruslan~R Salakhutdinov.
\newblock Reducing the dimensionality of data with neural networks.
\newblock {\em science}, 313(5786):504--507, 2006.

\bibitem{huang2017densely}
Gao Huang, Zhuang Liu, Laurens Van Der~Maaten, and Kilian~Q Weinberger.
\newblock Densely connected convolutional networks.
\newblock In {\em Proceedings of the IEEE conference on computer vision and pattern recognition}, pages 4700--4708, 2017.

\bibitem{jouppi2018motivation}
Norman Jouppi, Cliff Young, Nishant Patil, and David Patterson.
\newblock Motivation for and evaluation of the first tensor processing unit.
\newblock {\em ieee Micro}, 38(3):10--19, 2018.

\bibitem{kingma2013auto}
Diederik~P. Kingma and Max Welling.
\newblock Auto-encoding variational bayes.
\newblock In {\em International Conference on Learning Representations}, 2014.

\bibitem{cifar10}
Alex Krizhevsky.
\newblock Learning multiple layers of features from tiny images.
\newblock {\em University of Toronto}, 05 2012.

\bibitem{kullback1951information}
Solomon Kullback and Richard~A Leibler.
\newblock On information and sufficiency.
\newblock {\em The annals of mathematical statistics}, 22(1):79--86, 1951.

\bibitem{lecun2010mnist}
Yann LeCun, Corinna Cortes, and CJ~Burges.
\newblock Mnist handwritten digit database.
\newblock {\em ATT Labs [Online]. Available: http://yann.lecun.com/exdb/mnist}, 2, 2010.

\bibitem{li2022backdoor}
Yiming Li, Yong Jiang, Zhifeng Li, and Shu-Tao Xia.
\newblock Backdoor learning: A survey.
\newblock {\em IEEE Transactions on Neural Networks and Learning Systems}, 35(1):5--22, 2022.

\bibitem{2017magnet}
Dongyu Meng and Hao Chen.
\newblock Magnet: a two-pronged defense against adversarial examples, 2017.

\bibitem{oneto2020model}
Luca Oneto.
\newblock {\em Model selection and error estimation in a nutshell}.
\newblock Springer, 2020.

\bibitem{pajola2021fall}
Luca Pajola and Mauro Conti.
\newblock Fall of giants: How popular text-based mlaas fall against a simple evasion attack.
\newblock In {\em 2021 IEEE European Symposium on Security and Privacy (EuroS\&P)}, pages 198--211. IEEE, 2021.

\bibitem{paszke2019pytorch}
Adam Paszke, Sam Gross, Francisco Massa, Adam Lerer, James Bradbury, Gregory Chanan, Trevor Killeen, Zeming Lin, Natalia Gimelshein, Luca Antiga, Alban Desmaison, Andreas Kopf, Edward Yang, Zachary DeVito, Martin Raison, Alykhan Tejani, Sasank Chilamkurthy, Benoit Steiner, Lu~Fang, Junjie Bai, and Soumith Chintala.
\newblock Pytorch: An imperative style, high-performance deep learning library.
\newblock {\em arXiv preprint arXiv:1912.01703}, 2019.

\bibitem{shalev2014understanding}
Shai Shalev-Shwartz and Shai Ben-David.
\newblock {\em Understanding machine learning: From theory to algorithms}.
\newblock Cambridge university press, 2014.

\bibitem{shokri2017membership}
Reza Shokri, Marco Stronati, Congzheng Song, and Vitaly Shmatikov.
\newblock Membership inference attacks against machine learning models.
\newblock In {\em 2017 IEEE symposium on security and privacy (SP)}, pages 3--18. IEEE, 2017.

\bibitem{shumailov2021sponge}
Ilia Shumailov, Yiren Zhao, Daniel Bates, Nicolas Papernot, Robert Mullins, and Ross Anderson.
\newblock Sponge examples: Energy-latency attacks on neural networks.
\newblock In {\em 2021 IEEE European symposium on security and privacy (EuroS\&P)}, pages 212--231. IEEE, 2021.

\bibitem{sohn2015learning}
Kihyuk Sohn, Honglak Lee, and Xinchen Yan.
\newblock Learning structured output representation using deep conditional generative models.
\newblock {\em Advances in neural information processing systems}, 28, 2015.

\bibitem{tian2022comprehensive}
Zhiyi Tian, Lei Cui, Jie Liang, and Shui Yu.
\newblock A comprehensive survey on poisoning attacks and countermeasures in machine learning.
\newblock {\em ACM Computing Surveys}, 55(8):1--35, 2022.

\bibitem{PyTorchDenseNet}
Andreas Veit.
\newblock densenet-pytorch, 2018.
\newblock \\URL: \url{https://github.com/andreasveit/densenet-pytorch}.

\bibitem{speechcommandsdataset}
Pete Warden.
\newblock Speech commands: A dataset for limited-vocabulary speech recognition, 2018.

\bibitem{wolf2017why}
Marty~J Wolf, K~Miller, and Frances~S Grodzinsky.
\newblock Why we should have seen that coming: comments on microsoft's tay "experiment," and wider implications.
\newblock {\em SIGCAS Comput. Soc.}, 47(3):54–64, September 2017.

\bibitem{xiao2019seeing}
Qixue Xiao, Yufei Chen, Chao Shen, Yu~Chen, and Kang Li.
\newblock Seeing is not believing: Camouflage attacks on image scaling algorithms.
\newblock In {\em 28th USENIX Security Symposium (USENIX Security 19)}, pages 443--460, 2019.

\bibitem{xiao2018security}
Qixue Xiao, Kang Li, Deyue Zhang, and Weilin Xu.
\newblock Security risks in deep learning implementations.
\newblock In {\em 2018 IEEE Security and privacy workshops (SPW)}, pages 123--128. IEEE, 2018.

\end{thebibliography}


\end{document}